\documentclass[letterpaper, 10 pt, conference]{ieeeconf}  % Comment this line out if you need a4paper

\IEEEoverridecommandlockouts                              % This command is only needed if 
\usepackage{graphicx}
\usepackage{amsmath} % assumes amsmath package installed
\usepackage{amssymb}  % assumes amsmath package installed
\usepackage{subfigure}
\usepackage{algorithmic}
\usepackage{algorithm}
\usepackage{color}
\usepackage{hyperref}
\usepackage[bottom]{footmisc}

\title{\LARGE \bf
Sim-to-Real Transfer of Robotic Control with Dynamics Randomization
}

\author{Xue Bin Peng$^{1,2}$, Marcin Andrychowicz$^{1}$, Wojciech Zaremba$^{1}$, and Pieter Abbeel$^{1,2}$% <-this % stops a space
\thanks{$^{1}$OpenAI}%
\thanks{$^{2}$UC Berkeley, Department of Electrical Engineering and Computer Science}%
}

\begin{document}

\newcommand{\ch}[1]{\textcolor{blue}{#1}}

\maketitle
\thispagestyle{empty}
\pagestyle{empty}

%%%%%%%%%%%%%%%%%%%%%%%%%%%%%%%%%%%%%%%%%%%%%%%%%%%%%%%%%%%%%%%%%%%%%%%%%%%%%%%%
\begin{abstract}
Simulations are attractive environments for training agents as they provide an abundant source of data and alleviate certain safety concerns during the training process. But the behaviours developed by agents in simulation are often specific to the characteristics of the simulator. Due to modeling error, strategies that are successful in simulation may not transfer to their real world counterparts. In this paper, we demonstrate a simple method to bridge this ``reality gap''. By randomizing the dynamics of the simulator during training, we are able to develop policies that are capable of adapting to very different dynamics, including ones that differ significantly from the dynamics on which the policies were trained. This adaptivity enables the policies to generalize to the dynamics of the real world without any training on the physical system. Our approach is demonstrated on an object pushing task using a robotic arm. Despite being trained exclusively in simulation, our policies are able to maintain a similar level of performance when deployed on a real robot, reliably moving an object to a desired location from random initial configurations. We explore the impact of various design decisions and show that the resulting policies are robust to significant calibration error.\end{abstract}

%%%%%%%%%%%%%%%%%%%%%%%%%%%%%%%%%%%%%%%%%%%%%%%%%%%%%%%%%%%%%%%%%%%%%%%%%%%%%%%%
\section{INTRODUCTION}

Deep reinforcement learning (DeepRL) has been shown to be an effective framework for solving a rich repertoire of complex control problems. In simulated domains, agents have been developed to perform a diverse array of challenging tasks \cite{mnih-dqn-2015,LillicrapHPHETS15,DuanCHSA16}. Unfortunately, many of the capabilities demonstrated by simulated agents have often not been realized by their physical counterparts. Many of the modern DeepRL algorithms, which have spurred recent breakthroughs, pose high sample complexities, therefore often precluding their direct application to physical systems. In addition to sample complexity, deploying RL algorithms in the real world also raises a number of safety concerns both for the agent and its surroundings. Since exploration is a key component of the learning process, an agent can at times perform actions that endanger itself or its environment. Training agents in simulation is a promising approach that circumvents some of these obstacles. However, transferring policies from simulation to the real world entails challenges in bridging the "reality gap", the mismatch between the simulated and real world environments. Narrowing this gap has been a subject of intense interest in robotics, as it offers the potential of applying powerful algorithms that have so far been relegated to simulated domains. 

While significant efforts have been devoted to building higher fidelity simulators, we show that dynamics randomization using low fidelity simulations can also be an effective approach to develop policies that can be transferred directly to the real world. The effectiveness of our approach is demonstrated on an object pushing task, where a policy trained exclusively in simulation is able to successfully perform the task with a real robot without additional training on the physical system.

\begin{figure}[t]
	\centering
\includegraphics[width=1\columnwidth]{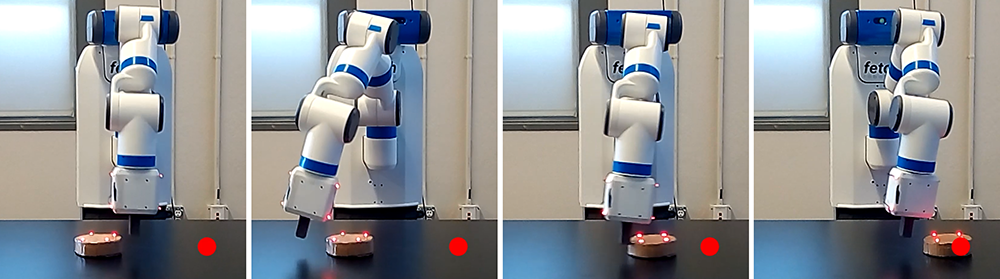}
	\caption{A recurrent neural network policy trained for a pushing task in simulation is deployed directly on a Fetch Robotics arm. The red marker indicates the target location for the puck.}
	\label{fig:screenshot0}
	\vspace{-0.5cm}
\end{figure}

\section{RELATED WORK}
Recent years have seen the application of deep reinforcement learning to a growing repertoire of control problems. The framework has enabled simulated agents to develop highly dynamic motor skills \cite{2016-TOG-deepRL,2017-TOG-deepLoco,Liu2017,HeessTSLMWTEWER17}. But due to the high sample complexity of RL algorithms and other physical limitations, many of the capabilities demonstrated in simulation have yet to be replicated in the physical world. Guided Policy Search (GPS) \cite{LevineWA15} represents one of the few algorithms capable of training policies directly on a real robot. By leveraging trajectory optimization with learned linear dynamics models, the method is able to develop complex manipulation skills with relatively few interactions with the environment. The method has also been extended to learning vision-based manipulation policies \cite{LevineFDA15}. Researchers have also explored parallelizing training across multiple robots \cite{LevinePKQ16}. Nonetheless, successful examples of training policies directly on physical robots have so far been demonstrated only on relatively restrictive domains.

\subsection{Domain Adaptation}
The problem of transferring control policies from simulation to the real world can be viewed as an instance of domain adaptation, where a model trained in a source domain is transfered to a new target domain. One of the key assumptions behind these methods is that the different domains share common characteristics such that representations and behaviours learned in one will prove useful for the other. Learning invariant features has emerged as a promising approach of taking advantage of these commonalities \cite{TzengDHFPLSD15,Ganin2016}. Tzeng et al. \cite{TzengDHFPLSD15} and Gupta et al. \cite{GuptaDLAL17} explored using pairwise constraints to encourage networks to learn similar embeddings for samples from different domains that are labeled as being similar. Daftry et al. \cite{DaftryBH16} applied a similar approach to transfer policies for controlling aerial vehicles to different environments and vehicle models. In the context of RL, adversarial losses have been used to transfer policies between different simulated domains, by encouraging agents to adopt similar behaviours across the various environments \cite{WulfmeierPA17}. Alternatively, progressive networks have also been used to transfer policies for a robotic arm from simulation to the real world \cite{RusuVRHPH16}. By reusing features learned in simulation, their method was able to significantly reduce the amount of data needed from the physical system. Christiano et al. \cite{ChristianoSMSBT16} transfered policies from simulation to a real robot by training an inverse-dynamics model from real world data. While promising, these methods nonetheless still require data from the target domain during training.

\subsection{Domain Randomization}
Domain randomization is a complementary class of techniques for adaptation that is particularly well suited for simulation. With domain randomization, discrepancies between the source and target domains are modeled as variability in the source domain. Randomization in the visual domain has been used to directly transfer vision-based policies from simulation to the real world without requiring real images during training \cite{SadeghiL16,TobinFRSZA17}. Sadeghi and Levine \cite{SadeghiL16} trained vision-based controllers for a quadrotor using only synthetically rendered scenes, and Tobin et al. \cite{TobinFRSZA17} demonstrated transferring image-based object detectors. Unlike previous methods, which sought to bridge the reality gap with high fidelity rendering \cite{JamesJ16}, their systems used only low fidelity rendering and modeled differences in visual appearance by randomizing scene properties such as lighting, textures, and camera placement. In addition to randomizing the visual features of a simulation, randomized dynamics have also been used to develop controllers that are robust to uncertainty in the dynamics of the system. Mordatch et al. \cite{MordatchLT15} used a trajectory optimizer to plan across an ensemble of dynamics models, to produce robust trajectories that are then executed on a real robot. Their method allowed a Darwin robot to perform a variety of locomotion skills. But due to the cost of the trajectory optimization step, the planning is performed offline. Other methods have also been proposed to develop robust policies through adversarial training schemes \cite{RajeswaranGLR16,PintoDSG17}. Yu et al. \cite{uposi2017} trained a system identification module to explicitly predict parameters of interest, such as mass and friction. The predicted parameters are then provided as input to a policy to compute the appropriate controls. While the results are encouraging, these methods have so far only been demonstrated on transfer between different simulators.

The work most reminiscent to our proposed method is that of Antonova et al. \cite{AntonovaCSK17}, where randomized dynamics was used to transfer manipulation policies from simulation to the real world. By randomizing physical parameters such as friction and latency, they were able to train policies in simulation for pivoting objects held by a gripper, and later transfer the policies directly to a Baxter robot without requiring additional fine-tuning on the physical system. However their policies were modeled using memoryless feedforward networks, and while the policies developed robust strategies, the lack of internal state limits the feedforward policies' ability to adapt to mismatch between the simulated and real environment. We show that memory-based policies are able to cope with greater variability during training and also better generalize to the dynamics of the real world. Unlike previous methods which often require meticulous calibration of the simulation to closely conform to the physical system, our policies are able to adapt to significant calibration error.

\subsection{Non-prehensile Manipulation}
Pushing, a form of non-prehensile manipulation, is an effective strategy for positioning and orienting objects that are too large or heavy to be grasped \cite{YuBFR16}. Though pushing has attracted much interest from the robotics community \cite{mason96,Dogar–2011,Fazeli2016}, it remains a challenging skill for robots to adopt. Part of the difficulty stems from accurately modeling the complex contact dynamics between surfaces. Characteristics such as friction can vary significantly across the surface of an object, and the resulting motions can be highly sensitive to the initial configuration of the contact surfaces \cite{YuBFR16}. Models have been proposed to facilitate planning algorithms \cite{mason96,Mason98,Dogar–2011}, but they tend to rely on simplifying assumptions that are often violated in practice. More recently, deep learning methods have been applied to train predictive models for pushing \cite{FinnGL16}. While data-driven methods overcome some of the modeling challenges faced by previous frameworks, they require a large corpus of real world data during training. Such a dataset can be costly to collect, and may become prohibitive for more complex tasks. Clavera et al. demonstrated transferring pushing policies trained in simulation to a real PR2 \cite{Clavera17}. Their approach took advantage of shaped reward functions and careful calibration to ensure that the behaviour of the simulation conforms to that of the physical system. In contrast, we will show that adaptive policies can be trained exclusively in simulation and using only sparse rewards. The resulting policies are able accommodate large calibration errors when deployed on a real robot and also generalize to variability in the dynamics of the physical system.

\section{BACKGROUND}
In this section we will provide a review of the RL framework and notation used in the following sections. We consider a standard RL problem where an agent interacts with an environment according to a policy in order to maximize a reward. The state of the environment at timestep $t$ is denoted by $s_t \in S$. For simplicity, we assume that the state is fully observable. A policy $\pi(a | s)$ defines a distribution over the action space $A$ given a particular state $s$, where each query to the policy samples an action $a$ from the conditional distribution. The reward function $r : S \times A \rightarrow \mathbb{R}$ provides a scalar signal that reflects the desirability of performing an action at a given state. For convenience, we denote $r_t = r(s_t, a_t)$. The goal of the agent is to maximize the multi-step return ${R_t = \sum_{t' = t}^T \gamma^{t' - t} r_{t'}}$, where $\gamma \in [0, 1]$ is a discount factor and $T$ is the horizon of each episode.

The objective during learning is to find an optimal policy $\pi^*$ that maximize the expected return of the agent $J(\pi)$
\[\pi^* = \mathop{\mathrm{arg \ max}}_\pi J(\pi) \]
If each episode starts in a fixed initial state, expected return can be rewritten as the expected return starting at the first step
\[J(\pi) = \mathbb{E}[R_0 | \pi] = \mathbb{E}_{\tau \sim p(\tau | \pi)} \left[ \sum_{t = 0}^{T - 1} r(s_t, a_t) \right] \]
where $p(\tau | \pi)$ represents the likelihood of a trajectory ${\tau = (s_0, a_0, s_1, ..., a_{T - 1}, s_T)}$ under the policy $\pi$,

\[p(\tau | \pi) = p(s_0) \prod_{t = 0}^{T-1} p(s_{t+1} | s_{t}, a_{t}) \pi(s_{t}, a_{t})\]
with the state transition model $p(s_{t + 1} | s_t, a_t)$ being determined by the dynamics of the environment. The dynamics is therefore of crucial importance, as it determines the consequences of the agent's actions, as well as the behaviours that can be realized.

\subsection{Policy Gradient Methods}
For a parametric policy $\pi_\theta$ with parameters $\theta$, the objective is to find the optimal parameters $\theta^*$ that maximizes the expected return $\theta^* = \mathop{\mathrm{arg \ max}}_\theta J(\pi_\theta)$. Policy gradient methods \cite{Sutton00policygradient} is a popular class of algorithms for learning parametric policies, where an estimate of the gradient of the objective $\triangledown_\theta J(\pi_\theta)$ is used to perform gradient ascent to maximize the expected return. While the previous definition of a policy is suitable for tasks where the goal is common across all episodes, it can be generalized to tasks where an agent is presented with a different goal every episode by constructing a universal policy \cite{schaul15}. A universal policy is a simple extension where the goal $g \in G$ is provided as an additional input to the policy $\pi(a | s, g)$. The reward is then also dispensed according to the goal $r(s_t, a_t, g)$. In our framework, a random goal will be sampled at the start of each episode, and held fixed over the course the episode. For the pushing task, the goal specifies the target location for an object.

\subsection{Hindsight Experience Replay}
During training, RL algorithms often benefit from carefully shaped reward functions that help guide the agent towards fulfilling the overall objective of a task. But designing a reward function can be challenging for more complex tasks, and may bias the policy towards adopting less optimal behaviours. An alternative is to use a binary reward $r(s, g)$ that only indicates if a goal is satisfied in a given state,
\[
r(s, g) =
\begin{cases}
0, & \text{if $g$ is satisfied in $s$}\\
-1, & \text{otherwise}
\end{cases}
\]
Learning from a sparse binary reward is known to be challenging for most modern RL algorithms. We will therefore leverage a recent innovation, Hindsight Experience Relay (HER) \cite{AndrychowiczWRS17}, to train policies using sparse rewards. Consider an episode with trajectory $\tau \in (s_0, a_0,...,a_{T-1}, s_T)$, where the goal $g$ was not satisfied over the course the trajectory. Since the goal was not satisfied, the reward will be $-1$ at every timestep, therefore providing the agent with little information on how to adjust its actions to procure more rewards. But suppose that we are provided with a mapping $m : S \rightarrow G$, that maps a state to the corresponding goal satisfied in the given state. For example, $m(s_T) = g'$ represents the goal that is satisfied in the final state of the trajectory. Once a new goal has been determined, rewards can be recomputed for the original trajectory under the new goal $g'$. While the trajectory was unsuccessful under the original goal, it becomes a successful trajectory under the new goal. Therefore, the rewards computed with respect to $g'$ will not be $-1$ for every timestep. By replaying past experiences with HER, the agent can be trained with more successful examples than is available in the original recorded trajectories. So far, we have only considered replaying goals from the final state of a trajectory. But HER is also amenable to other replay strategies, and we refer interested readers to the original paper \cite{AndrychowiczWRS17} for more details.

\section{METHOD}

Our objective is to train policies that can perform a task under the dynamics of the real world $p^*(s_{t+1} | s_t, a_t)$. Since sampling from the real world dynamics can be prohibitive, we instead train a policy using an approximate dynamics model $\hat{p}(s_{t+1} | s_t, a_t) \approx p^*(s_{t+1} | s_t, a_t)$ that is easier to sample from. For all of our experiments, $\hat{p}$ assumes the form of a physics simulation. Due to modeling and other forms of calibration error, behaviours that successfully accomplish a task in simulation may not be successful once deployed in the real world. Furthermore, it has been observed that DeepRL policies are prone to exploiting idiosyncrasies of the simulator to realize behaviours that are infeasible in the real world \cite{LillicrapHPHETS15,HeessTSLMWTEWER17}. Therefore, instead of training a policy under one particular dynamics model, we train a policy that can perform a task under a variety of different dynamics models. First we introduce a set of dynamics parameters $\mu$ that parameterizes the dynamics of the simulation $\hat{p}(s_{t+1} | s_t, a_t, \mu)$.  The objective is then modified to maximize the expected return across a distribution of dynamics models $\rho_\mu$,

\[\mathop{\mathbb{E}}_{\mu \sim \rho_\mu} \left[\mathbb{E}_{\tau \sim p(\tau | \pi, \mu)} \left[ \sum_{t = 0}^{T - 1} r(s_t, a_t) \right]  \right]\]
By training policies to adapt to variability in the dynamics of the environment, the resulting policy might then better generalize to the dynamics of real world.

\subsection{Tasks}
\begin{figure}[t]
	\centering
\subfigure{\includegraphics[width=0.4\columnwidth]{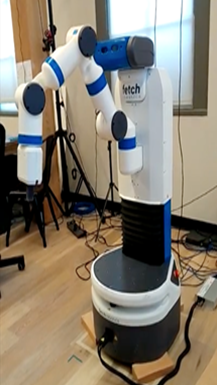}}
    \hspace{0.05\columnwidth}
    \subfigure{\includegraphics[width=0.4\columnwidth]{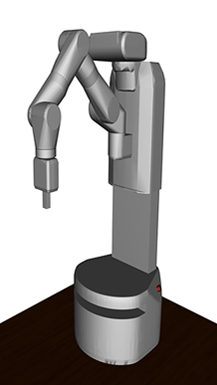}}
    \vspace{-0.2cm}
	\caption{Our experiments are conducted on a 7-DOF Fetch Robotics arm. \textbf{Left:} Real robot. \textbf{Right:} Simulated MuJoCo model.}
	\label{fig:fetch}
    \vspace{-0.5cm}
\end{figure}

Our experiments are conducted on a puck pushing task using a 7-DOF Fetch Robotics arm. Images of the real robot and simulated model is available in Figure \ref{fig:fetch}. The goal $g$ for each episode specifies a random target position on the table that the puck should be moved to. The reward is binary with $r_t = 0$ if the puck is within a given distance of the target, and $r_t = -1$ otherwise. At the start of each episode, the arm is initialized to a default pose and the initial location of the puck is randomly placed within a fixed bound on the table.

\subsection{State and Action}
The state is represented using the joint positions and velocities of the arm, the position of the gripper, as well as the puck's position, orientation, linear and angular velocities. The combined features result in a 52D state space. Actions from the policy specify target joint angles for a position controller. Target angles are specified as relative offsets from the current joint rotations. This yields a 7D action space.

\subsection{Dynamics Randomization}
During training, rollouts are organized into episodes of a fixed length. At the start of each episode, a random set of dynamics parameters $\mu$ are sampled according to $\rho_\mu$ and held fixed for the duration of the episode. The parameters which we randomize include:
\begin{itemize}
  \item Mass of each link in the robot's body
  \item Damping of each joint
  \item Mass, friction, and damping of the puck
  \item Height of the table
  \item Gains for the position controller
  \item Timestep between actions
  \item Observation noise
\end{itemize}
which results in a total of 95 randomized parameters. The timestep between actions specifies the amount of time an action is applied before the policy is queried again to sample a new action. This serves as a simple model of the latency exhibited by the physical controller. The observation noise models uncertainty in the sensors and is implemented as independent Gaussian noise applied to each state feature. While parameters such as mass and damping are constant over the course of an episode, the action timestep and the observation noise varies randomly each timestep.

\subsection{Adaptive Policy}
Manipulation tasks, such as pushing, have a strong dependency on the physical properties of the system (e.g. mass, friction, and characteristics of the actuators). In order to determine the appropriate actions, a policy requires some means of inferring the underlying dynamics of its environment. While the dynamics parameters are readily available in simulation, the same does not hold once a policy has been deployed in the real world. In the absence of direct knowledge of the parameters, the dynamics can be inferred from a history of past states and actions. System identification using a history of past trajectories has been previously explored by Yu et al. \cite{uposi2017}. Their system incorporates an online system identification module $\phi(s_t, h_t) = \hat{\mu}$, which utilizes a history of past states and actions $h_t = [a_{t-1}, s_{t-1}, a_{t-2}, s_{t-2}, ...]$ to predict the dynamics parameters $\mu$. The predicted parameters are then used as inputs to a universal policy that samples an action according to the current state and inferred dynamics $\pi(a_t | s_t, \hat{\mu})$. However, this decomposition requires identifying the dynamics parameters of interest to be predicted at runtime, which may be difficult for more complex systems. Constructing such a set of parameters necessarily requires some structural assumptions about the dynamics of a system, which may not hold in the real world. Alternatively, SysID can be implicitly embedded into a policy by using a recurrent model $\pi(a_t | s_t, z_t, g)$, where the internal memory $z_t = z(h_t)$ acts as a summary of past states and actions, thereby providing a mechanism with which the policy can use to infer the dynamics of the system. This model can then be trained end-to-end and the representation of the internal memory can be learned without requiring manual identification of a set of dynamics parameters to be inferred at runtime.

\subsection{Recurrent Deterministic Policy Gradient}
Since HER augments the original training data recorded from rollouts of the policy with additional data generated from replayed goals, it requires off-policy learning. Deep Deterministic Policy Gradient (DDPG) \cite{LillicrapHPHETS15} is a popular off-policy algorithm for continuous control. Its extension to recurrent policies, Recurrent Deterministic Policy Gradient (RDPG) \cite{HeessHLS15}, provides a method to train recurrent policies with off-policy data. To apply RDPG, we denote a deterministic policy as $\pi(s_t, z_t, g) = a_t$. In additional to the policy, we will also model a recurrent universal value function as $Q(s_t, a_t, y_t, g, \mu)$, where $y_t = y(h_t)$ is the value function's internal memory. Since the value function is used only during training and the dynamics parameters $\mu$ of the simulator are known, $\mu$ is provided as an additional input to the value function but not to the policy. We will refer to a value function with knowledge of the dynamics parameters as an \emph{omniscient critic}. This follows the approach of \cite{FoersterAFW16a,LoweWTHAM17}, where additional information is provided to the value function during training in order to reduce the variance of the policy gradients and allow the value function to provide more meaningful feedback for improving the policy.

Algorithm \ref{alg:training} summarizes the training procedure, where $M$ represents a replay buffer \cite{LillicrapHPHETS15}, and $\theta$ and $\varphi$ are the parameters for the policy and value function respectively. We also incorporate target networks \cite{LillicrapHPHETS15}, but they are excluded for brevity.

\begin{algorithm}[h!]
    \caption{Dynamics Randomization with HER and RDPG}
    \label{alg:training}
    \begin{algorithmic}[1]
    \STATE{$\theta \leftarrow$ random weights}
    \STATE{$\varphi \leftarrow$ random weights}

    \WHILE{not done}
    \STATE{$g \sim \rho_g$ sample goal}
    \STATE{$\mu \sim \rho_\mu$ sample dynamics}
    \STATE{Generate rollout $\tau=(s_0, a_0,...,s_T)$with dynamics $\mu$}
	\FOR{each $s_t, a_t$ in $\tau$}
		\STATE{$r_t \leftarrow r(s_t, g)$}
	\ENDFOR
    \STATE{Store $(\tau, \{r_t\}, g, \mu)$ in $M$}
    
    \STATE{Sample episode $(\tau, \{r_t\}, g, \mu)$ from $M$}
    \STATE{\textbf{with} probability $k$}
        \STATE{$\quad g \leftarrow$ replay new goal with HER}
    	\STATE{$\quad r_t \leftarrow r(s_t, g)$ for each $t$}
    \STATE{\textbf{endwith}}
    \item[]
    
    \FOR{each $t$}
    	\STATE{Compute memories $z_t$ and $y_t$}
    	\STATE{$\hat{a}_{t+1} \leftarrow \pi_\theta(s_{t+1}, z_{t+1}, g)$}
        \STATE{$\hat{a}_t \leftarrow \pi_\theta(s_t, z_t, g)$}
    	\STATE{$q_t \leftarrow r_t + \gamma Q_\varphi(s_{t+1}, \hat{a}_{t+1}, y_{t+1}, g, \mu)$}
        \STATE{$\triangle q_t \leftarrow q_t - Q_\varphi(s_t, a_t, y_t, g, \mu)$}
    \ENDFOR
    
    \STATE{${\triangledown_\varphi = \frac{1}{T} \sum_t \triangle q_t \frac{\partial Q_\varphi(s_t, a_t, y_t, g, \mu)}{\partial \varphi}}$}
    \STATE{${\triangledown_\theta = \frac{1}{T} \sum_t \frac{\partial Q_\varphi(s_t, \hat{a}_t, y_t, g, \mu)}{\partial a} \frac{\partial \hat{a}_t}{\partial \theta}}$}
    
    \STATE{Update value function and policy with $\triangledown_\theta$ and $\triangledown_\varphi$}
\ENDWHILE
\end{algorithmic}
\end{algorithm}

\subsection{Network Architecture}

\begin{figure}[t]
	\vspace{0.3cm}
	\centering
	\subfigure{\includegraphics[width=0.8\columnwidth]{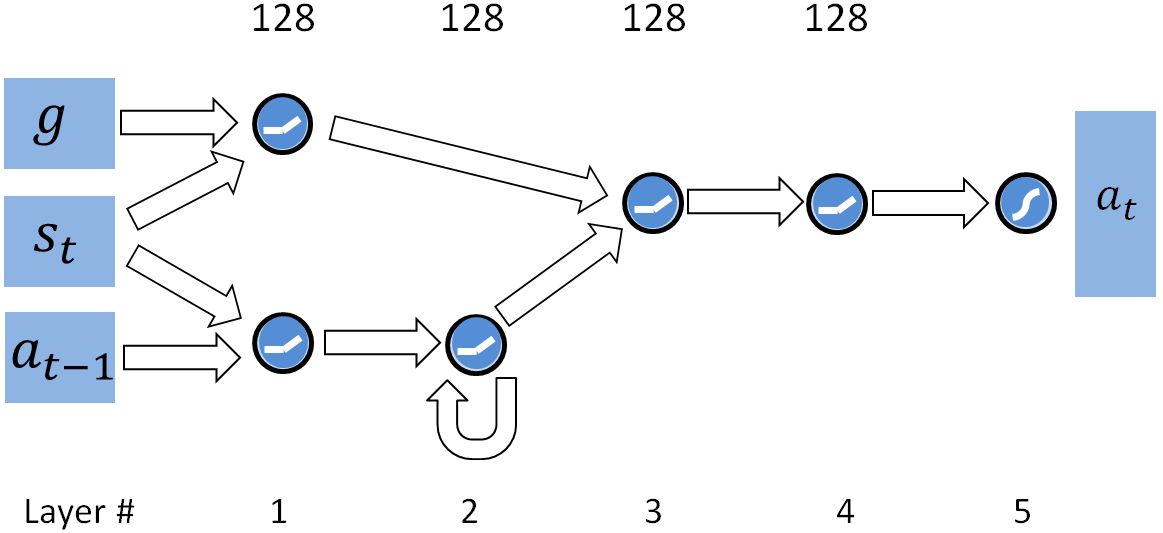}}
    \subfigure{\includegraphics[width=0.8\columnwidth]{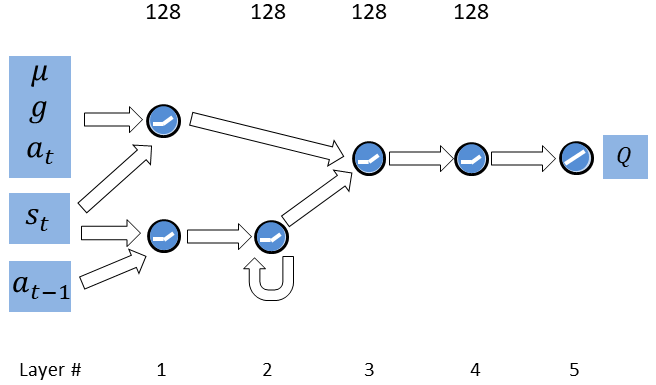}}
	\caption{Schematic illustrations of the policy network \textbf{(top)}, and value network \textbf{(bottom)}. Features that are relevant for inferring the dynamics of the environment are processed by the recurrent branch, while the other inputs are processed by the feedforward branch.}
	\label{fig:poliNet}
    \vspace{-0.75cm}
\end{figure}

A schematic illustrations of the policy and value networks are available in Figure \ref{fig:poliNet}. The inputs to the network consist of the current state $s_t$ and previous action $a_{t-1}$, and the internal memory is updated incrementally at every step. Each network consists of a feedforward branch and recurrent branch, with the latter being tasked with inferring the dynamics from past observations. The internal memory is modeled using a layer of LSTM units and is provided only with information required to infer the dynamics (e.g. $s_t$ and $a_{t-1}$). The recurrent branch consists of an embedding layer of 128 fully-connected units followed by 128 LSTM units. The goal $g$ does not hold any information regarding the dynamics of the system, and is therefore processed only by the feedforward branch. Furthermore, since the current state $s_t$ is of particular importance for determining the appropriate action for the current timestep, a copy is also provided as input to the feedforward branch. This presents subsequent layers with more direct access to the current state, without requiring information to filter through the LSTM. The features computed by both branches are then concatenated and processed by 2 additional fully-connected layers of 128 units each. The value network $Q(s_t, a_t, a_{t - 1}, g, \mu)$ follows a similar architecture, with the query action $a_t$ and parameters $\mu$ being processed by the feedforward branch. ReLU activations are used after each hidden layer (apart from the LSTM). The output layer of $Q$ consists of linear units, while $\pi$ consists of tanh output units scaled to span the bounds of each action parameter.

\begin{figure}[t]
	\vspace{0.2cm}
	\centering
\subfigure{\includegraphics[width=1\columnwidth]{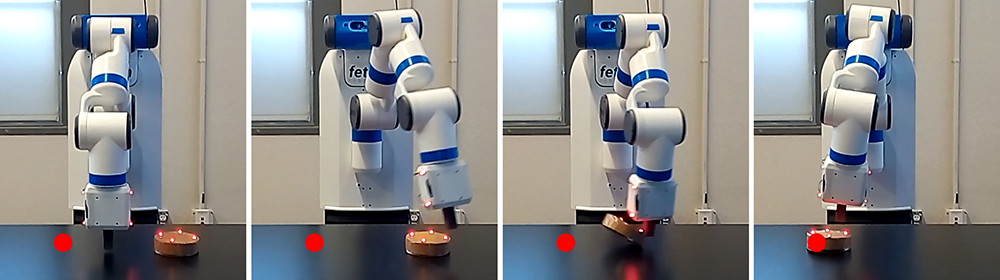}}
\subfigure{\includegraphics[width=1\columnwidth]{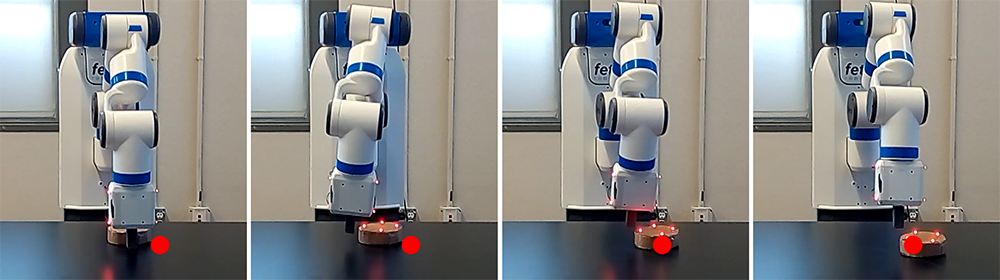}}
\subfigure{\includegraphics[width=1\columnwidth]{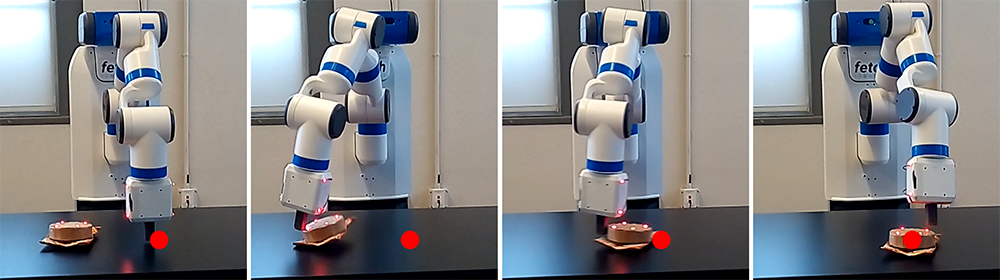}}
	\caption{LSTM policy deployed on the Fetch arm. \textbf{Bottom:} The contact dynamics of the puck was modified by attaching a packet of chips to the bottom.}
	\label{fig:screenshots1}
    \vspace{-0.5cm}
\end{figure}

\begin{table}[b]
\vspace{-0.4cm}
{ \centering  
\begin{tabular}{|l|c|}
\hline
{\bf Parameter} & {\bf Range} \\ \hline
Link Mass &  $[0.25, 4] \times$ default mass of each link \\ \hline
Joint Damping &  $[0.2, 20] \times$ default damping of each joint \\ \hline
Puck Mass &  $[0.1, 0.4] kg$ \\ \hline
Puck Friction &  $[0.1, 5]$ \\ \hline
Puck Damping &  $[0.01, 0.2] Ns/m$ \\ \hline
Table Height &  $[0.73, 0.77] m$ \\ \hline
Controller Gains &  $[0.5, 2] \times$ default gains \\ \hline
Action Timestep $\lambda$ &  $[125, 1000]s^{-1}$ \\ \hline
\end{tabular} \\
}
\caption{Dynamics parameters and their respective ranges.}
\label{tab:paramVals}
\vspace{-0.5cm}
\end{table}

\section{EXPERIMENTS}
Results are best seen in the supplemental video \href{https://youtu.be/XUW0cnvqbwM}{https://youtu.be/XUW0cnvqbwM}. Snapshots of policies deployed on the real robot are available in Figure \ref{fig:screenshots1}. All simulations are performed using the MuJoCo physics engine \cite{TodorovET12} with a simulation timestep of 0.002s. 20 simulation timesteps are performed for every control timestep. Each episode consists of 100 control timestep, corresponding to approximately 4 seconds per episode, but may vary as a result of the random timesteps between actions. Table \ref{tab:paramVals} details the range of values for each dynamics parameter. At the start of each episode, a new set of parameters $\mu$ is sampled by drawing values for each parameter from their respective range. Parameters such as mass, damping, friction, and controller gains are logarithmically sampled, while other parameters are uniformly sampled. The timestep $\triangle t$ between actions varies every step according to $\triangle t \sim \triangle t_0 + \mathrm{Exp}(\lambda)$,
where $\triangle t_0 = 0.04s$ is the default control timestep, and $\mathrm{Exp}(\lambda)$ is an exponential distribution with rate parameter $\lambda$. While $\triangle t$ varies every timestep, $\lambda$ is fixed within each episode. In addition to randomizing the physical properties of the simulated environment, we also simulate sensor noise by applying gaussian noise to the observed state features at every step. The noise has a mean of zero and a standard deviation of $5\%$ of the running standard deviation of each feature. Gaussian action exploration noise is added at every step with a standard deviation of $0.01 rad$.

\begin{figure}[b]
	\vspace{-0.5cm}
	\centering
	\includegraphics[width=0.95\columnwidth]{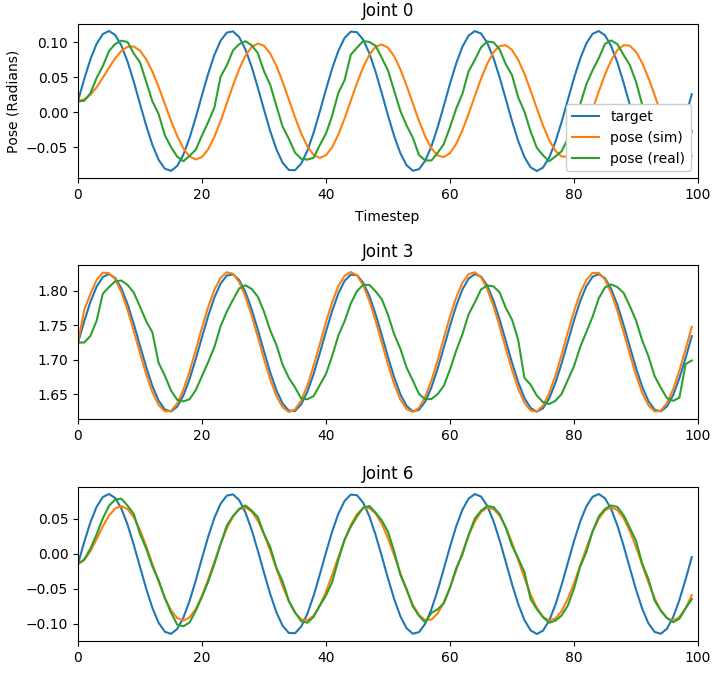}
	\vspace{-0.2cm}
	\caption{Joint trajectories recorded from the simulated and real robot when executing the same target trajectories. The joints correspond to the shoulder, elbow, and wrist of the Fetch arm.}
	\label{fig:trajs}
\end{figure}

The real puck has a mass of approximately $0.2kg$ and a radius of 0.065m. The goal is considered satisfied if the puck is within 0.07m of the target. The location of the puck is tracked using the PhaseSpace mocap system. When evaluating performance on the physical system, each episode consists of 200 timesteps. Little calibration was performed to ensure that the behaviour of the simulation closely conforms to that of the real robot. While more extensive calibration will likely improve performance, we show that our policy is nonetheless able to adapt to the physical system despite poor calibration. To illustrate the discrepancies between the dynamics of the real world and simulation we executed the same target trajectory on the real and simulated robot, and recorded the resulting joint trajectories. Figure \ref{fig:trajs} illustrates the recorded trajectories. Given the same target trajectory, the pose trajectories of the simulated and real robot differ significantly, with varying degrees of mismatch across joints.

During training, parameter updates are performed using the ADAM optimizer \cite{KingmaB14} with a stepsize of $5\times10^{-4}$ for both the policy and value function. Updates are performed using batches of 128 episodes with 100 steps per episode. New goals are sampled using HER with a probability of $k=0.8$. Each policy is trained for approximately 8000 update iterations using about 100 million samples, which requires approximately 8 hours to simulate on a 100 core cluster.

\begin{figure}[b]
	\vspace{-0.5cm}
	\centering
	\includegraphics[width=0.9\columnwidth]{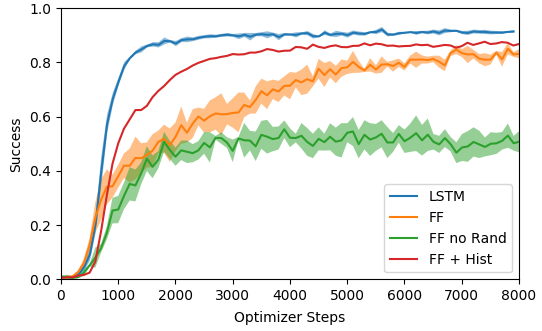}
    \vspace{-0.35cm}
	\caption{Learning curves of different network architectures. Four policies are trained for each architecture with different random seeds. Performance is evaluated over 100 episodes in simulation with random dynamics.}
	\label{fig:learningCurves}
    \vspace{-0.2cm}
\end{figure}

\subsection{Comparison of Architectures}
To evaluate the impact of different architectural choices, we compared policies modeled using different architectures and tested their performance in simulation and on the real robot. The first is an LSTM policy following the architecture illustrated in Figure \ref{fig:poliNet}. Next we consider a memoryless feedforward network (FF) that receives only the current state $s_t$ and goal $g$ as input. As a baseline, we also trained a memoryless feedforward network without randomization \mbox{(FF no Rand)}, then evaluated the performance with randomization. To provide the feedforward network with more information to infer the dynamics, we augmented the inputs with a history of the 8 previously observed states and actions (FF + Hist). The success rate is determined as the portion of episodes where the goal is fulfilled at the end of the episode. In simulation, performance of each policy is evaluated over 100 episodes, with randomized dynamics parameters for each episode. Learning curves comparing the performance of different model architectures in simulation are available in Figure \ref{fig:learningCurves}. Four policies initialized with different random seeds are trained for each architecture. The LSTM learns faster while also converging to a higher success rate than the feedforward models. The feedforward network trained without randomization is unable to cope with unfamiliar dynamics during evaluation. While training a memoryless policy with randomization improves robustness to random dynamics, it is still unable to perform the task consistently.

\begin{figure}[t]
\vspace{0.2cm}
	\centering
	\includegraphics[width=0.8\columnwidth]{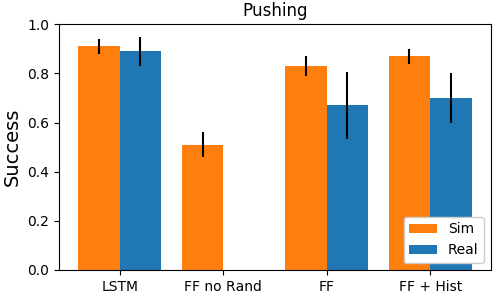}
    \vspace{-0.25cm}
	\caption{Performance of different models when deployed on the simulated and real robot for the pushing task. Policies are trained using only data from simulation.}
	\label{fig:perfReal}
    \vspace{-0.5cm}
\end{figure}

Next, we evaluate the performance of the different models when deployed on the real Fetch arm. Figure \ref{fig:perfReal} compares the performance of the final policies when deployed in simulation and the real world. Table \ref{tab:perf} summarizes the performance of the models. The target and initial location of the puck is randomly placed within a $0.3m \times 0.3m$ bound. While the performance of LSTM and FF + Hist policies are comparable in simulation, the LSTM is able to better generalize to the dynamics of the physical system. The feedforward network trained without randomization is unable to perform the task under the real world dynamics.

\subsection{Ablation}
To evaluate the effects of randomizing the various dynamics parameters, we trained policies with subsets of the parameters held fixed. A complete list of the dynamics parameters are available in Table \ref{tab:paramVals}.
The configurations we consider include training with a fixed timestep between actions, training without observation noise, or with fixed mass for each link. Table \ref{tab:ablationPerf} summarizes the performance of the resulting policies when deployed on the real robot. Disabling randomization of the action timestep, observation noise, link mass, and friction impairs the policies' ability to adapt to the physical environment. Policies trained without randomizing the action timestep and observation noise show particularly noticeable drops in performance. This suggests that coping with the latency of the controller and sensor noise are important factors in adapting to the physical system.

\subsection{Robustness}
To evaluate the robustness of the LSTM policy to different dynamics when deployed on the real robot, we experimented with changing the contact dynamics of the physical system by attaching a packet of chips to the bottom of the puck. The texture of the bag reduces the friction between the puck and the table, while the contents of the bag further alters the contact dynamics. Nonetheless, the LSTM policy achieves a success rate of $0.91 \pm 0.04$, which is comparable to the success rate without the attachment $0.89 \pm 0.06$. The policy also develops clever strategies to make fine adjustments to position the puck over the target. One such strategy involves pressing on one side of the puck in order to partially upend it before sliding it to the target. Other strategies including manipulating the puck from the top or sides depending on the required adjustments, and correcting for case where the puck overshoots the target. These behaviours emerged naturally from the learning process using only a sparse binary reward.

\begin{table}[t]
{ \centering  
\vspace{0.3cm}
\begin{tabular}{|l|c|c|c|}
\hline
{\bf Model} & {\bf Success (Sim)} & {\bf Success (Real)} & {\bf Trials (Real)} \\ \hline
LSTM & $\boldsymbol{0.91 \pm 0.03}$ & $\boldsymbol{0.89 \pm 0.06}$ & $28$ \\ \hline
FF no Rand &  $0.51 \pm 0.05$ & $0.0 \pm 0.0$ & $10$ \\ \hline
FF &  $0.83 \pm 0.04$ & $0.67 \pm 0.14$ & $12$ \\ \hline
FF + Hist &  $0.87 \pm 0.03$ & $0.70 \pm 0.10$ & $20$ \\ \hline
\end{tabular} \\
}
\caption{Performance of the policies when deployed on the simulated and real robot. Performance in simulation is evaluated over 100 trials with randomized dynamics parameters.}
\label{tab:perf}
\vspace{-0.4cm}
\end{table}

\begin{table}[t]
{ \centering  
\begin{tabular}{|l|c|c|c|}
\hline
{\bf Model} & {\bf Success} & {\bf Trials} \\ \hline
all & $\boldsymbol{0.89 \pm 0.06}$ & $28$ \\ \hline
fixed action timestep & $0.29 \pm 0.11$ & $17$ \\ \hline
no observation noise &  $0.25 \pm 0.12$ & $12$ \\ \hline
fixed link mass &  $0.64 \pm 0.10$ & $22$ \\ \hline
fixed puck friction & $0.48 \pm 0.10$ & $27$ \\ \hline
\end{tabular} \\
}
\caption{Performance of LSTM policies on the real robot, where the policies are trained with subsets of parameters held fixed.}
\label{tab:ablationPerf}
\vspace{-0.75cm}
\end{table}

\section{Conclusions}
We demonstrated the use of dynamics randomization to train recurrent policies that are capable of adapting to unfamiliar dynamics at runtime. Training policies with randomized dynamics in simulation enables the resulting policies to be deployed directly on a physical robot despite poor calibrations. By training exclusively in simulation, we are able to leverage simulators to generate a large volume of training data, thereby enabling us to use powerful RL techniques that are not yet feasible to apply directly on a physical system. Our experiments with a real world pushing tasks showed comparable performance to simulation and the ability to adapt to changes in contact dynamics. We also evaluated the importance of design decisions pertaining to choices of architecture and parameters which to randomize during training. We intend to extend this work to a richer repertoire tasks and incorporate more modalities such as vision. We hope this approach will open more opportunities for developing skillful agents in simulation that are then able to be deployed in the physical world.

\section{Acknowledgement}
We would like to thank Ankur Handa, Vikash Kumar, Bob McGrew, Matthias Plappert, Alex Ray, Jonas Schneider, and Peter Welinder for their support and feedback on this project.

\bibliographystyle{ieeetran}
\bibliography{dorand}

\end{document}